\documentclass[10pt,twocolumn,letterpaper]{article}

\usepackage[pagenumbers]{cvpr} 
\usepackage{booktabs,multirow,makecell,graphicx,array}
\usepackage[table]{xcolor}

\usepackage{algorithm}
\usepackage{algorithmic}
\usepackage{wrapfig}
\definecolor{cvprblue}{rgb}{0.21,0.49,0.74}
\usepackage[pagebackref,breaklinks,colorlinks,allcolors=cvprblue]{hyperref}

\def\paperID{9921} 
\def\confName{CVPR}
\def\confYear{2026}

\title{Rethinking Cross-Generator Image Forgery Detection through DINOv3}

\author{
     Zhenglin Huang$^{1}$    \quad Jason Li $^{2}$  \quad Haiquan Wen$^{1}$ \quad Tianxiao Li$^{1}$ \quad   Xi Yang$^{3}$ \\ \quad Lu Qi$^{4}$ \quad Bei Peng$^{5}$  \quad Xiaowei Huang$^{1}$  \quad Ming-Hsuan Yang$^{4}$ \quad Guangliang Cheng$^{1}$ \textsuperscript{$\dagger$}  \vspace{0.3em} \\
     {\normalsize $^1$University of Liverpool, UK \quad $^2$Nanyang Technological University } \\
     {\normalsize $^3$ HKUST \quad $^4$ UC Merced \quad $^5$ University of Sheffield  } \\
    {\normalsize Github Page: \url{https://github.com/hzlsaber/FGTS/}} \\
    {\normalsize \textsuperscript{$\dagger$} Corresponding author. E-mail: guangliang.cheng@liverpool.ac.uk }
}

\begin{document}
\maketitle
\begin{abstract}
As generative models become increasingly diverse and powerful, cross-generator detection has emerged as a new challenge.
Existing detection methods often memorize artifacts of specific synthesis models rather than learning transferable cues, leading to substantial failures on unseen generators. 
%
Thus, the detection models must have strong generalization to meet task-specific adaptation. 
Surprisingly, in this work, we find that frozen visual foundation models—particularly DINOv3—already act as good cross-generator detectors, achieving strong performance even \textbf{without} any fine-tuning.
Through systematic studies across frequency, spatial, and token perspectives, we identify representational trends, suggesting that DINOv3 relies on global, low-frequency structures.
These features serve as weak yet transferable \textbf{authenticity cues}, rather than on high-frequency, generator-specific artifacts. 
Motivated by these, we introduce a simple yet generalizable training-free token-ranking strategy followed by a lightweight linear probe.
We select only a small subset of authenticity-relevant tokens that consistently improve detection accuracy across all evaluated datasets.
Our study provides empirical evidence and a feasible hypothesis for why foundation models generalize across diverse generators.
As a result, our model builds a universal, efficient, and interpretable baseline for image forgery detection.

\end{abstract}
    
\section{Introduction}
\label{sec:intro}

The rapid evolution of generative AI has fundamentally reshaped visual content creation. 
High-fidelity images synthesized by diffusion and adversarial models~\cite{DBLP:journals/corr/abs-2406-14555, DBLP:journals/corr/abs-2503-04641,DBLP:conf/nips/DhariwalN21,DBLP:journals/corr/abs-2106-15282} are now virtually indistinguishable from real photographs, 
posing significant challenges to visual authenticity verification. 
Despite extensive progress, existing detectors still exhibit poor cross-generator generalization: they perform well on generators seen during training but fail catastrophically on unseen ones. 
This persistent gap raises a central question in modern forgery detection: 
\textbf{\textit{How can we develop a detector that generalizes across unseen generators?}}

\begin{figure}[t]
  \centering
   \includegraphics[width=1.0\linewidth]{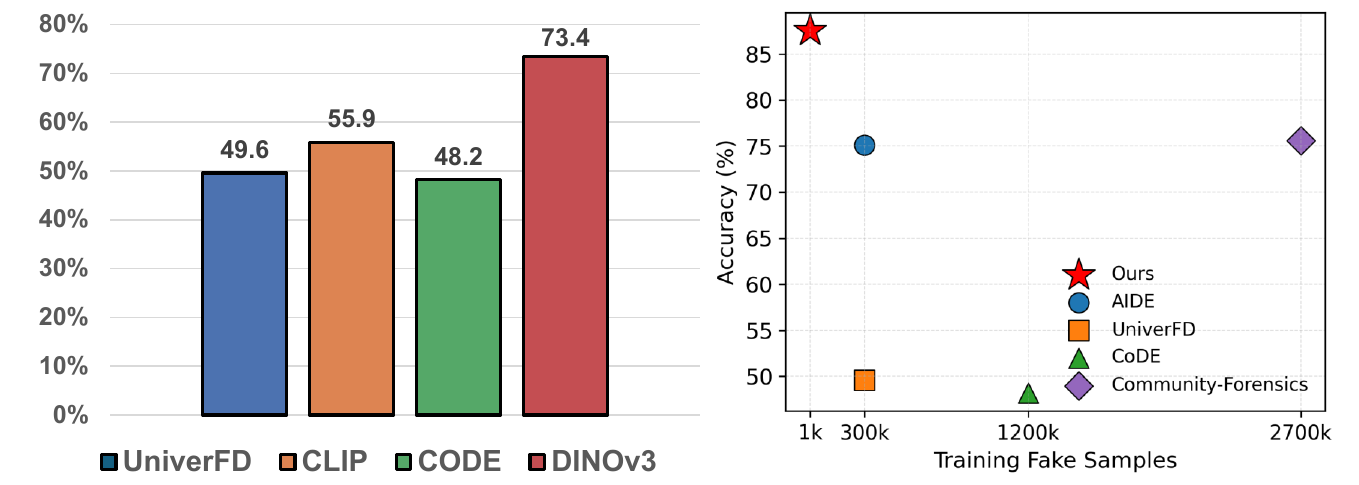}
\caption{
\textbf{Cross-generator performance comparison.} 
\textbf{Left:} Average OOD accuracy across the ten commercial generators in So-Fake-OOD~\cite{DBLP:journals/corr/abs-2505-18660}. 
\textbf{Right:} OOD detection accuracy versus the amount of training fake data for foundation model-based approaches. 
}

   \vspace{-1em}
   \label{figure:1}
\end{figure}

Recent advances in generalized image forgery detection have increasingly focused on model-centric approaches, leveraging powerful pre-trained backbones to improve cross-generator generalization~\cite{DBLP:conf/cvpr/OjhaLL23,DBLP:conf/eccv/BaraldiCCBNC24,DBLP:conf/aaai/LinLLYL25,DBLP:conf/cvpr/CozzolinoPCNV22,DBLP:conf/iclr/YanLCHJ0X25}. 
However, most of these methods remain heavily reliant on large-scale, task-specific adaptation. 
UniverFD~\cite{DBLP:conf/cvpr/OjhaLL23} fine-tunes a CLIP backbone~\cite{DBLP:conf/icml/RadfordKHRGASAM21} on more than 300k samples, 
and CoDE~\cite{DBLP:conf/eccv/BaraldiCCBNC24} trains an entirely new contrastive space using millions of diffusion-generated images. 
As shown in Fig.~\ref{figure:1} (left), both methods suffer substantial degradation on unseen commercial generators in the So-Fake-OOD benchmark~\cite{DBLP:journals/corr/abs-2505-18660}. 

This phenomenon raises a fundamental question: {\it \textbf{Does the large-scale adaptation process itself introduce bias, or are the foundation models inherently limited?}}
To investigate this, we compare UniverFD directly against its frozen foundation backbone (CLIP). 
Remarkably, as shown in Fig.~\ref{figure:1} (left), the frozen CLIP model achieves \textbf{55.9\%} accuracy on So-Fake-OOD, outperforming its adapted counterpart by \textbf{6.3\%}.
This counter-intuitive finding suggests a critical limitation: large-scale adaptation with hundreds of thousands of generated images is not only unnecessary, but may also bias the model's intrinsic generalization capability by overfitting to the artifacts of specific generators.

Although the frozen CLIP model performs better than its adapted counterpart, its overall detection accuracy remains limited. 
We posit that this limitation stems from its training objective, which aligns visual and linguistic semantics rather than modeling purely visual regularities~\cite{DBLP:conf/icml/RadfordKHRGASAM21}. 
Consequently, such language-guided supervision may overlook visual cues that distinguish real from generated content.

Building on this hypothesis, we next explore whether a foundation model trained purely on visual objectives can inherently generalize better for this task. 
To this end, we turn to \textbf{DINOv3}~\cite{DBLP:journals/corr/abs-2508-10104}, the latest strong baseline in vision foundation models trained exclusively on large-scale image data without any language supervision. 
Unlike CLIP, DINOv3 learns from image-only self-distillation, encouraging invariance to transformations while preserving fine-grained visual regularities.
Under the same zero shot protocol, the frozen DINOv3 model achieves an average accuracy of \textbf{73.4\%} on the So-Fake-OOD benchmark (Fig.~\ref{figure:1} (left)), substantially surpassing all the task-specific detectors trained with hundreds of thousands of samples.
This performance is remarkable since DINOv3 was never trained for authenticity discrimination or any forgery-related objective.

These results naturally raise another central question:
\textbf{\textit{Why does a frozen, vision-only foundation model generalize so well to unseen generators?}}
To explore this, we analyze DINOv3’s internal representations (Sec.~\ref{sec:analysis}) across three dimensions—frequency, spatial structure, and token organization.
Our findings converge to a consistent pattern: DINOv3 encodes a low-frequency, globally coherent yet spatially distributed signal that systematically differentiates real images from generated ones. 
We refer to this emergent pattern as \textbf{authenticity cues}, a form of visual regularity that is not tied to any specific generator and that is unevenly distributed across patch tokens rather than concentrated in 
summary tokens such as CLS or register tokens.

Building on these findings, we introduce \textbf{Fisher-Guided Token Selection (FGTS)}, 
a training-free token ranking framework (Sec.~\ref{sec:method}) that identifies the patch 
tokens most strongly associated with authenticity cues. 
To implement this ranking, FGTS adopts the Fisher score~\cite{fisher1936,gu2012generalized}, 
which measures each token’s real/fake separability by comparing between-class differences 
with within-class variability. Using this measure, we find that a compact subset of patch 
tokens preserves intrinsic authenticity signals while reducing generator-specific noise.

While FGTS uncovers meaningful separability in the frozen feature space, we further explore whether a small amount of supervision can enhance these compact representations without resorting to large-scale adaptation. 
To this end, we employ a lightweight linear probe trained on only \textbf{1k} fake images from a 
single generator—two orders of magnitude fewer than prior methods—which preserves the 
frozen model’s generalization while achieving state-of-the-art cross-generator performance.

\noindent Our main contributions are summarized as follows:
\begin{itemize}[leftmargin=0.5cm, itemindent=0cm]
    \item \textbf{Empirical Finding.} We uncover that the frozen visual foundation model \textbf{DINOv3} exhibits strong cross-generator generalization, surpassing all large-scale training-based detectors under a training-free protocol.

    \item \textbf{Interpretive Analysis.} Through systematic observations, we identify consistent representational trends suggesting that DINOv3 may encode \textbf{authenticity cues} through globally coherent, low-frequency structures rather than local artifacts.

    \item \textbf{Methodological Insight.} 
    Building on these observations, we propose \textbf{Fisher-Guided Token Selection (FGTS)}, 
    a simple yet effective framework that isolates authenticity-relevant patch tokens from frozen representations. Using only 1k training fake images, FGTS attains \textbf{87.5\%} accuracy on So-Fake-OOD (Fig.~\ref{figure:1} (right)) and \textbf{92.6\%} on GenImage~\cite{DBLP:conf/nips/ZhuCYHLLT0H023}, demonstrating that compact token-level representations can capture strong cross-generator signals with minimal supervision.

\end{itemize}

\section{Preliminaries}
\label{sec:preliminaries}

\subsection{Problem Formulation}
\label{subsec:problem}

Cross-generator image forgery detection aims to determine whether an image $x$ is real or AI-generated. We define a binary classifier $f_\theta: \mathcal{I} \to \{\text{real}, \text{fake}\}$, where $\mathcal{I}$ denotes the image space. Images are drawn from either the real distribution $p_{\text{real}}$ or a generator-specific fake distribution $p_{\text{fake}}^{(g)}$, where $g \in \mathcal{G}$ is a generative model:
\begin{equation}
x \sim
\begin{cases}
p_{\text{real}}(x), & x \in \mathcal{I}_{\text{real}}, \\
p_{\text{fake}}^{(g)}(x), & x \in \mathcal{I}_{\text{fake}}^{(g)},\; g \in \mathcal{G}.
\end{cases}
\label{eq:distribution}
\end{equation}

In the cross-generator setting, the detector is trained on images from $\mathcal{G}_{\text{seen}}$ and evaluated on disjoint unseen generators $\mathcal{G}_{\text{unseen}}$, where $\mathcal{G}_{\text{seen}} \cap \mathcal{G}_{\text{unseen}} = \varnothing$. The objective is to learn $f_\theta$ that generalizes across the distributional shift between generators in $\mathcal{G}_{\text{seen}}$ and $\mathcal{G}_{\text{unseen}}$.

\subsection{Representational Concepts}
\label{subsec:concepts}

To characterize the types of information that may influence cross-generator generalization, we analyze the visual cues captured by Vision Transformers~\cite{DBLP:conf/iclr/DosovitskiyB0WZ21} along two independent but complementary dimensions.

\noindent \textbf{Frequency Dimension.}
Visual information can be analyzed in the frequency domain:
\begin{itemize}[leftmargin=0.5cm]
\item \textbf{Low-frequency (LF) components}: encode global structure, smooth color or luminance transitions, overall layout, and lighting coherence. These signals often remain stable across generation paradigms~\cite{DBLP:conf/icml/FrankESFKH20,DBLP:conf/nips/DzanicSW20}.
\item \textbf{High-frequency (HF) components}: encode local details, sharp edges, and fine textures, where generator-specific artifacts frequently manifest, such as checkerboard patterns in GANs~\cite{DBLP:journals/corr/AitkenLTCWS17,DBLP:conf/cvpr/KarrasLAHLA20} or synthesis noise characteristics~\cite{DBLP:conf/icassp/CorviCZPNV23}.
\end{itemize}

\noindent \textbf{Spatial Dimension.}
Visual signals can also be characterized by spatial scope:
\begin{itemize}[leftmargin=0.5cm]
\item \textbf{Local patterns}: spatially confined content within patches or small neighborhoods, including localized inconsistencies~\cite{DBLP:conf/iccv/RosslerCVRTN19}.
\item \textbf{Global structure}: image-level coherence requiring integration across distant regions, including scene layout, long-range dependencies, and perspective or illumination consistency~\cite{Nightingale2022AIsynthesizedFA,DBLP:conf/icip/JuJKXNL22,DBLP:conf/cvpr/DurallKK20}.
\end{itemize}

Together, these two dimensions describe the primary forms of visual information that a model may exploit to distinguish real and generated images.
In Sec.~\ref{subsec:freq_spatial}, we empirically examine how frozen foundation models, particularly DINOv3, respond to perturbations along these dimensions, providing insight into the representational basis of cross-generator generalization.

\subsection{Foundation Models}
\label{subsec:foundation_models}

Frozen DINOv3 exhibits strong cross-generator generalization compared to adapted vision-language models (Figure~\ref{figure:1}). 
We summarize the relevant paradigms and highlight the representational 
factors that will be examined in Sec.~\ref{sec:analysis}.

\noindent \textbf{CLIP: Vision-Language Alignment.}
CLIP (Contrastive Language-Image Pre-training)~\cite{DBLP:conf/icml/RadfordKHRGASAM21} aligns visual and textual representations through contrastive learning between paired image and text data. 
This objective encourages semantic alignment across modalities and emphasizes category-level correspondence rather than fine-grained visual regularities.

\noindent \textbf{DINO: Vision-Only Self-Distillation.}
The DINO family~\cite{DBLP:conf/iccv/CaronTMJMBJ21,DBLP:journals/tmlr/OquabDMVSKFHMEA24,DBLP:journals/corr/abs-2508-10104} learns visual representations purely from images through self-distillation, where a student network matches a teacher’s predictions across multiple augmented views to promote invariance while preserving discriminative structure.
In this work, we focus on \textbf{DINOv3}, a vision-only self-distilled transformer whose architecture includes three token types: CLS, register, and spatial patch tokens. These play distinct functional roles, as 
summarized below.

\textbf{CLS Token.} A global summarizer trained purely on visual consistency without linguistic supervision as in CLIP.

\textbf{Register Tokens.} Learnable non-spatial tokens designed to store image-level statistics and prevent artifact accumulation in patch tokens~\cite{DBLP:conf/iclr/DarcetOMB24,DBLP:journals/corr/abs-2506-08010}. 
In DINOv2~\cite{DBLP:journals/tmlr/OquabDMVSKFHMEA24}, these tokens can be attached post hoc after training, whereas DINOv3~\cite{DBLP:journals/corr/abs-2508-10104} integrates them natively during training, allowing all token types to be learned jointly from the beginning.

\textbf{Patch Tokens.} Spatially grounded tokens that represent localized image content and serve as the primary carriers of visual detail.

This architecture provides multiple token types that capture visual information at different levels of abstraction. 
In Sec.~\ref{subsec:token_analysis}, we empirically analyze how these representations behave across spatial, frequency, and token-level dimensions to better understand their contribution to cross-generator robustness.

\section{Empirical Observations and Analysis}
\label{sec:analysis}

Building on the frequency–spatial framework introduced in Sec.~\ref{sec:preliminaries}, we conduct a series of empirical analyses to uncover how DINOv3 encodes authenticity-related information.
Specifically, Sec.~\ref{subsec:freq_spatial} studies DINOv3 from two complementary 
perspectives—the \textbf{frequency domain} and \textbf{spatial perturbations}. 
Sec.~\ref{subsec:token_analysis} then investigates the \textbf{token-level mechanisms} that 
underpin global image representations. 
Together, these observations motivate the hypothesis and method developed in Sec.~\ref{sec:method}.

To ensure a fair and representative evaluation of cross-generator generalization, our analyses focus on the \textbf{commercial diffusion generators} in the So-Fake-OOD benchmark~\cite{DBLP:journals/corr/abs-2505-18660}, which encompass ten major proprietary text-to-image systems with diverse architectures and training pipelines. 
This subset best reflects real-world generative diversity and unseen conditions. Comprehensive quantitative 
results on other generators, including GANs and open-source diffusion models, are reported in Sec.~\ref{sec:exp}.

\subsection{Frequency-Spatial Evidence}
\label{subsec:freq_spatial}

Prior work has shown that different generative families tend to introduce artifacts 
with distinct spectral and structural characteristics. 
For example, GANs often produce high-frequency patterns such as checkerboard effects 
\cite{DBLP:conf/icml/FrankESFKH20,DBLP:conf/nips/DzanicSW20}, 
while diffusion models exhibit more subtle low-frequency deviations in color or geometry 
\cite{DBLP:conf/icassp/CorviCZPNV23}.  
These findings suggest that frequency and spatial organization provide a natural lens 
for studying how authenticity-related signals may appear across different generators.  
Motivated by this perspective, we examine which components along these two dimensions 
DINOv3 is actually sensitive to when generalizing to unseen models.
This consideration leads to two guiding questions.  
First, does DINOv3 rely more on low-frequency structure or on high-frequency details?  
Second, are the relevant cues primarily local or globally coherent?  
The following two empirical observations address these questions.

\begin{figure}[t]
  \centering
  \includegraphics[width=1.0\linewidth]{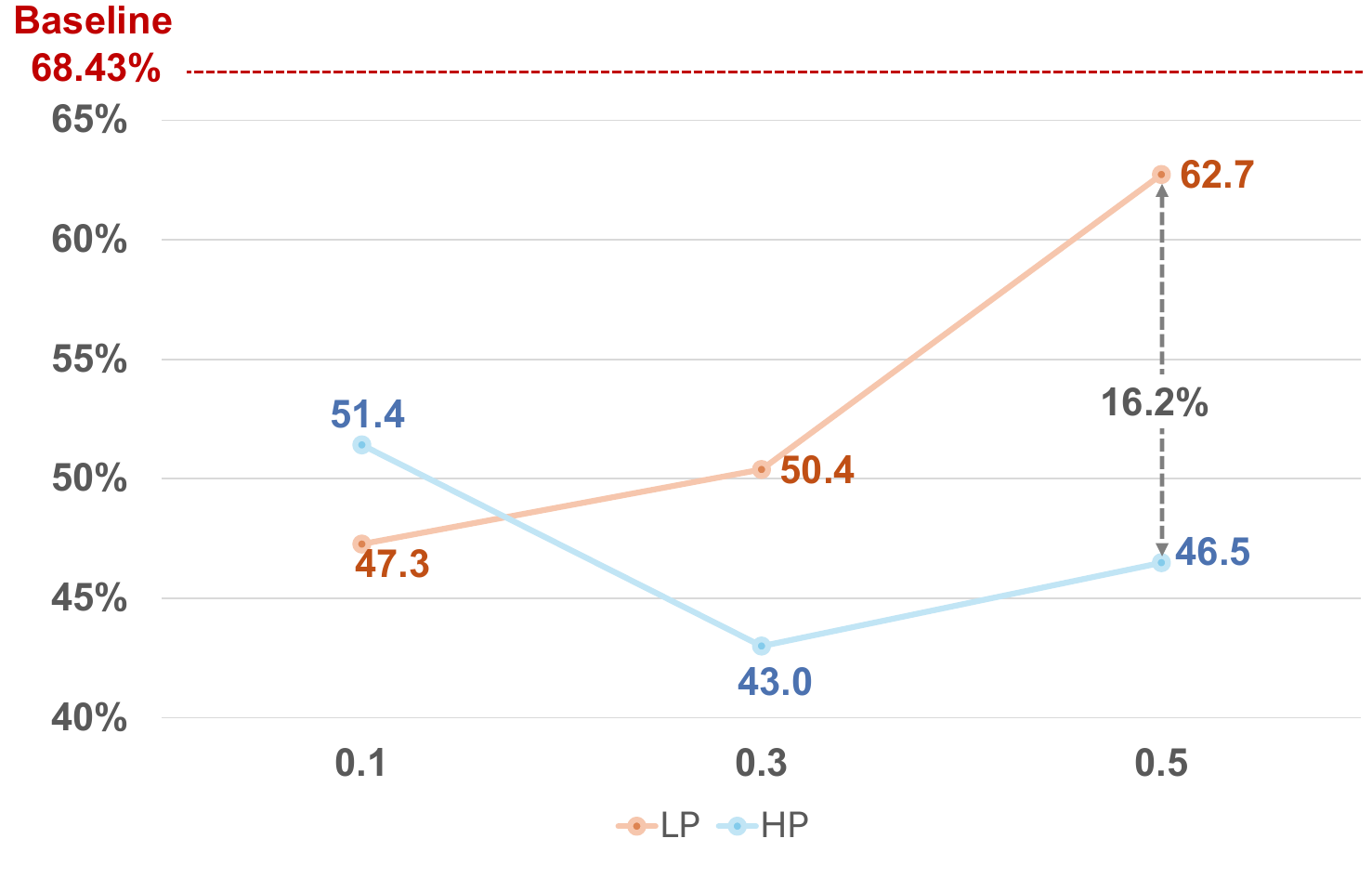}
  \caption{\textbf{Low-pass vs.\ high-pass filtering on DINOv3.}
  Average accuracy under low-pass (LP) and high-pass (HP) filtering across cutoff ratios on ten diffusion generators.}
  \vspace{-5mm}
  \label{figure:freq_curve}
\end{figure}

\noindent\textbf{Observation A: Low-frequency dominance.}  
We first probe DINOv3’s spectral sensitivity by applying ideal frequency domain filters with cutoff ratios $r\!\in\!\{0.1,0.3,0.5\}$. 
Under the \textbf{low-pass (LP)} condition, only coarse, low-frequency components are preserved, while fine details are removed.
The \textbf{high-pass (HP)} condition keeps high-frequency signals while discarding global structure. 
Average accuracy across ten commercial diffusion generators (Fig.~\ref{figure:freq_curve}) reveals a pronounced asymmetry: As the cutoff increases, LP accuracy steadily recovers, whereas HP accuracy remains almost unchanged and far below the baseline.  
At $r\!=\!0.5$, the LP–HP gap reaches \textbf{16.2\%}, confirming that DINOv3’s authenticity representation is strongly biased toward low-frequency information.  
This tendency likely stems from its self-distillation objective, which promotes invariance to high-frequency variations and emphasizes stable, slowly varying spatial statistics.

\noindent\textbf{Observation B: Dependence on global spatial coherence.}
While frequency analysis reveals a strong preference for low-frequency information, it remains unclear whether these cues originate from locally smooth textures or truly global structural organization.
To clarify this, we apply two complementary spatial perturbations:
(1) \textbf{Random Masking}, which removes local content by replacing 50\% of patches with their mean pixel values while preserving the global layout; and
(2) \textbf{Patch Shuffling}, which destroys spatial coherence by randomly permuting patch positions within local neighborhoods while preserving local textures.

\begin{figure}[t]
  \centering
  \includegraphics[width=1.0\linewidth]{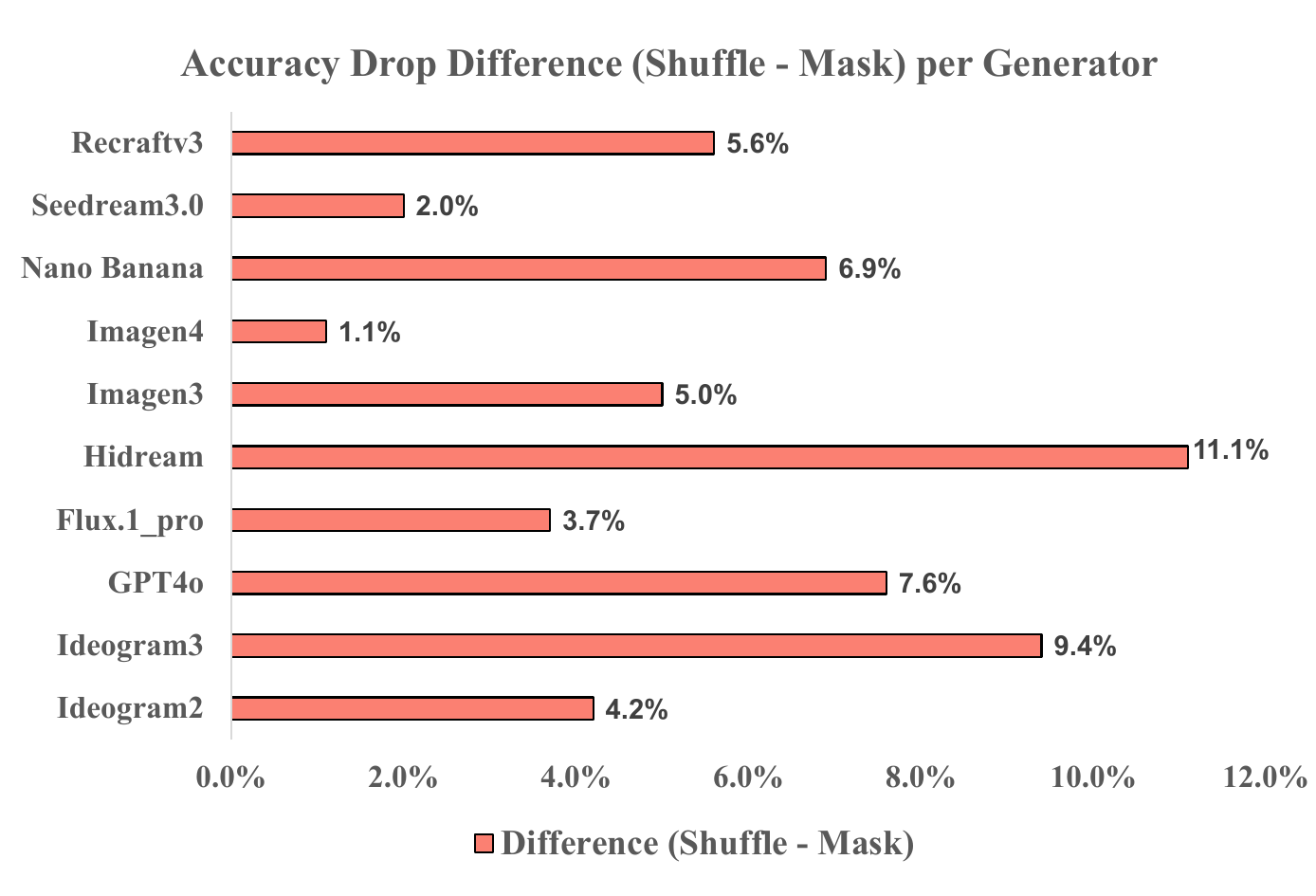}
  \caption{
  \textbf{Impact of spatial coherence on DINOv3.}
  Accuracy drop difference ($\Delta$Acc, Shuffle–Mask) under 50\% perturbation across ten commercial diffusion generators from So-Fake-OOD.}
  \label{figure:mask_shuffle}
  \vspace{-5mm}
  \end{figure}

Fig.~\ref{figure:mask_shuffle} visualizes the \textit{accuracy drop difference} (Shuffle–Mask) across ten commercial diffusion generators.
DINOv3 shows near invariance to masking (average drop \textbf{0.1\%}), confirming that local pixel removal has minimal effect.
However, once spatial coherence is disrupted, performance decreases sharply (average drop \textbf{5.6\%}, up to \textbf{11.1\%} on HiDream).
This result demonstrates that DINOv3’s decision boundary critically depends on maintaining global spatial organization rather than on localized details. 

However, because frequency and spatial structure are coupled in natural images, these observations still leave open whether the low-frequency preference arises from retained spectral energy or coherent spatial organization.

\begin{figure*}[ht]
    \centering
    \includegraphics[width=\linewidth]{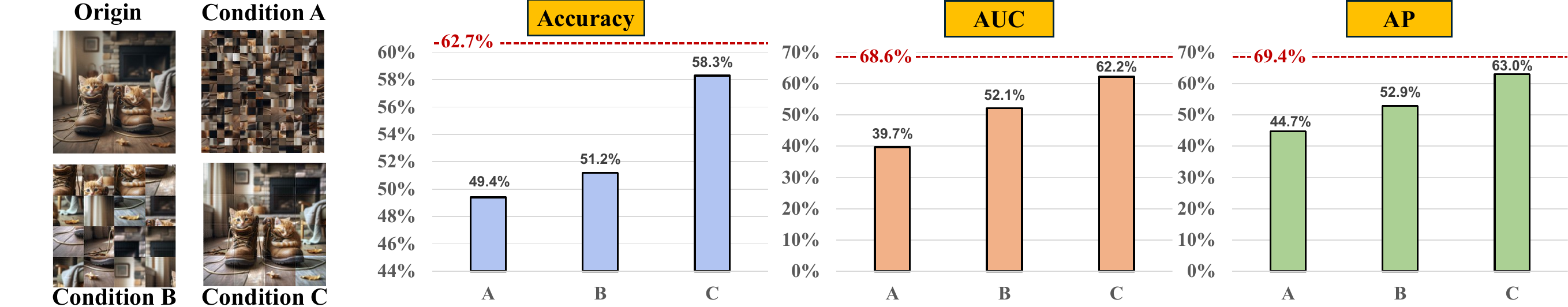}
    \caption{
        \textbf{Performance under spatial disruption conditions.} 
        Left: Visual examples of each condition. Right: Accuracy, AUC, and AP across three experimental conditions. The red dashed line indicates the low-pass only baseline (LP, $r{=}0.5$) without spatial disruption. 
    }
    \label{fig:spatial-coherence}
\end{figure*}

\noindent\textbf{Observation C: Frequency and spatial interaction.}  
Building on Observations A and B, we next examine whether the low-frequency preference is driven by preserved spatial structure rather than by retained spectral energy alone.  
To separate these factors, we conduct a joint frequency and spatial perturbation analysis under three controlled conditions (Fig.~\ref{fig:spatial-coherence}).  
\textbf{Condition A} applies a global low-pass filter followed by complete patch shuffling, removing all global structural coherence.  
\textbf{Condition B} uses block-wise low pass filtering with intra block shuffling, disrupting coarse spatial alignment while preserving local textures.  
\textbf{Condition C} applies the same block-wise low pass filtering without shuffling as a control.

Across our setting, all metrics exhibit a clear and consistent trend.  
\textbf{Condition A} collapses to near random behavior, indicating that the discriminative signal vanishes once global spatial structure is destroyed.  
\textbf{Condition B} yields only partial recovery, suggesting that isolated local structure provides limited support.  
\textbf{Condition C} stays close to the low pass baseline, confirming that block-wise low pass filtering alone introduces minimal artifacts.  
Overall, these results demonstrate that DINOv3’s apparent low-frequency dependence is in fact a dependence on spatially coherent global structure rather than on low-frequency energy itself.

\noindent\textbf{Interpretation.}  
Together, the three observations reveal a consistent representational pattern.  
DINOv3 distinguishes real from generated images through globally coherent low-frequency structures that encode spatial organization and overall scene coherence.  
Performance remains stable when local details are removed but collapses once global coherence is disrupted, indicating that structural organization rather than local texture drives authenticity perception.  
This global structural bias explains DINOv3’s strong generalization, since inconsistencies in perspective, illumination, and layout tend to persist across different diffusion models, whereas local artifacts vary widely.  
However, these findings describe the phenomenon at the image level and leave open a key question: 
\textbf{\textit{How is this global authenticity information encoded in the model, and which tokens contribute to it?}}

\subsection{Where Are Authenticity Cues Encoded?}
\label{subsec:token_analysis}

The previous section showed that DINOv3 distinguishes real from synthetic images by relying on \textbf{globally coherent low-frequency structures}.  
We refer to these structures as \textbf{authenticity cues}, a form of generator agnostic signal that reflects the physical and perceptual consistency of an image.  
Having characterized these cues at the image level, we next ask how they are represented inside DINOv3.  
Since the architecture combines spatial patch tokens with non-spatial CLS and register tokens 
(Sec.~\ref{subsec:foundation_models}), a key question is whether authenticity cues reside primarily in the non-spatial tokens or emerge as a distributed pattern across patches.
To address this, we perform controlled perturbation analyses and discriminability evaluations on each token type.

\begin{figure}[t]
  \centering
  \includegraphics[width=\linewidth]{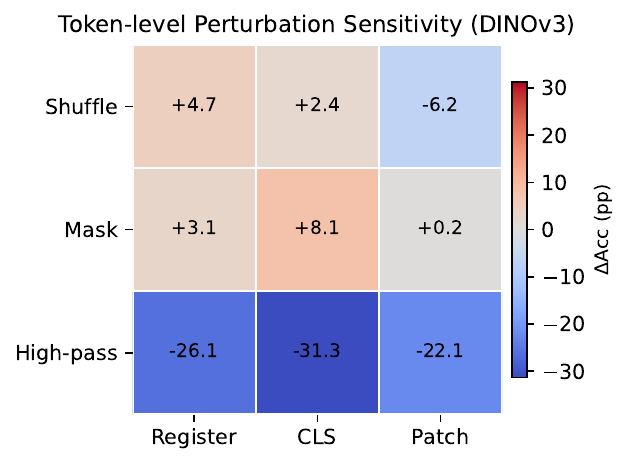}
\caption{\textbf{Token-level perturbation sensitivity in DINOv3.}
Heatmap of $\Delta\mathrm{Acc}$ (percentage points), where
$\Delta\mathrm{Acc} = \mathrm{Acc}_{\text{pert}} - \mathrm{Acc}_{\text{base}}$ for each token type 
under Shuffle, Mask, and High-pass perturbations.  
Positive values (red) indicate increased accuracy and negative values (blue) 
indicate decreased accuracy.}

  \label{figure:token_perturb}
  \vspace{-5mm}
\end{figure}

\noindent\textbf{Experimental Protocol.}
To examine the role of different token types, we extend the perturbation analyses from Sec.~\ref{subsec:freq_spatial} to the token level.  
For each token type (CLS, Register, Patch) in DINOv3's final layer, we  
(1) extract features under frequency and spatial perturbations,  
(2) compute the mean feature of real and synthetic samples for that token type using a small, balanced reference set, and  
(3) evaluate detection performance by comparing test samples against these mean features.  
This provides a training-free measure of how well each token type separates real/fake images.

To further assess the discriminative contribution of each token type, we additionally train a lightweight linear classifier on top of features from each token subset using 1,000 real and 1,000 synthetic images from a single generator.  
This complementary evaluation reveals how linearly separable each token type is under minimal supervision.

\begin{table}[t]
\centering
\small
\setlength{\tabcolsep}{5pt}
\renewcommand{\arraystretch}{1.1}
\caption{\textbf{Token-wise evaluation on frozen DINOv3.}
Average accuracy (Acc), AUC, and AP across evaluation sets under a linear probe protocol. 
Patch-only nearly matches using all tokens, while CLS and Register tokens lag behind, 
indicating that authenticity information is distributed among spatial patches.}
\label{table:tokens}
\vspace{1mm}
\resizebox{\linewidth}{!}{
\begin{tabular}{lccc}
\toprule
\textbf{Token strategy} & \textbf{Acc} & \textbf{AUC} & \textbf{AP} \\
\midrule
All (201 tokens)         & 0.7364 & 0.8100 & 0.8228 \\
CLS (1 token)            & 0.7053 & 0.8059 & 0.8199 \\
REG (4 tokens)           & 0.6827 & 0.7703 & 0.7785 \\
Patch (196 tokens)       & \textbf{0.7403} & \textbf{0.8132} & \textbf{0.8262} \\
CLS + REG (5 tokens)   & 0.7134 & 0.8002 & 0.8020 \\
CLS + Patch (197 tokens) & 0.7346 & 0.8070 & 0.8208 \\
\bottomrule
\end{tabular}}
\vspace{-3mm}
\end{table}

\noindent\textbf{Observation D: Distributed representation of global authenticity.}  
Fig.~\ref{figure:token_perturb} reveals distinct behaviors across token types.  
Non-spatial tokens (CLS and Register) are largely position independent, showing improved accuracy under shuffling ($+2.4\%, +4.7\%$) and substantial degradation under high-pass filtering ($-31.3\%, -26.1\%$).  
This pattern indicates that these tokens rely primarily on low-frequency global statistics rather than on spatial structure.  
Patch tokens show the opposite trend.  
Their accuracy decreases when spatial order is disrupted ($-6.2\%$), confirming their dependence on coherent spatial organization, 
yet they exhibit comparable low-frequency sensitivity under high-pass filtering ($-22.1\%$).  
Tab.~\ref{table:tokens} further shows that patch-only features achieve the highest accuracy (74.0\%), surpassing CLS (70.5\%) and Register (68.3\%) tokens.  

\noindent\textbf{Interpretation.}
These results suggest that DINOv3 distributes global information across both non-spatial and spatial tokens, but with different functional roles.
CLS and Register tokens capture broad, position-agnostic low-frequency statistics, whereas patch tokens integrate these low-frequency components with coherent spatial organization.
Since authenticity cues hinge on globally coherent low-frequency structure (Sec.~\ref{subsec:token_analysis}), such spatial grounding becomes particularly important for real/fake discrimination.
This explains why patch-only features (74.0\%) slightly outperform using all tokens (73.6\%): patch tokens carry the most task-relevant portion of DINOv3’s global low-frequency representation.

\section{Method}
\label{sec:method}

\begin{table*}[t]
\small
\centering
\setlength{\tabcolsep}{3pt}
\renewcommand{\arraystretch}{1}
\caption{\textbf{Cross-generator generalization on So-Fake-OOD.}
All numbers report accuracy (\%). 
SD and PG denote models fine-tuned on SD~v1.4 and ProGAN, respectively, 
and LD denotes the CLIP adapted using LDM-generated images.}
\resizebox{\linewidth}{!}{
\begin{tabular}{cccccccccccc}
\toprule
\textbf{Detection method} &
Flux.1\_pro &
GPT-4o &
HiDream &
Ideogram~2 &
Ideogram~3 &
Imagen~3 &
Imagen~4 &
Recraft-v3 &
Seedream~3.0 &
Nano~Banana &
\textbf{Avg-acc} \\ 
\midrule

Community-Forensics~\cite{DBLP:conf/cvpr/ParkO25}  & 59.37 & 86.14 & 80.48 & \textbf{66.53} & 77.68 & 75.36 & 77.74 & \underline{75.65} & 58.59 & 81.87 & \underline{75.61} \\
UniverFD~\cite{DBLP:conf/cvpr/OjhaLL23}    & 49.58 & 49.63 & 65.43 & 45.29 & 46.95 & 45.96 & 47.15 & 47.93 & 49.66 & 48.18 & 49.57 \\
CLIP-LD~\cite{DBLP:conf/icml/RadfordKHRGASAM21}   &66.80  &69.67  &71.13  &53.15 &66.69  &77.42  &67.06  &73.71  &68.38  &74.14 & 68.82 \\
CoDE~\cite{DBLP:conf/eccv/BaraldiCCBNC24}      & 48.71 & 42.36 & 41.03 & 43.62 & 45.18 & 46.03 & 45.88 & 49.66 & 46.74 & 44.39 & 45.37 \\
AIDE-PG~\cite{DBLP:conf/iclr/YanLCHJ0X25} &49.91 &\textbf{97.42} &56.46 & 51.31 & \underline{95.71}& \textbf{94.71} & \underline{77.89}   &71.91 &59.83 &\textbf{96.12}  &75.13  \\

AIDE-SD~\cite{DBLP:conf/iclr/YanLCHJ0X25} &51.15 &88.55 & 63.21 & 53.15 & 90.31 & 86.22 &72.81 &68.29 &\underline{70.71} &92.61 & 73.70 \\
\midrule

Ours (Training free)     & \underline{68.90} &84.20 &\underline{83.33} &56.96 &83.55 &79.59 &69.90 &71.00 &67.42 &85.77 &75.06 \\

Ours (Linear probe)    & \textbf{79.90} & \underline{96.04} & \textbf{96.02} & \underline{63.80} & \textbf{95.91} & \underline{92.83} & \textbf{90.53} & \textbf{86.57} & \textbf{77.87} & \underline{95.88} & \textbf{87.53} \\
\bottomrule

\end{tabular}}

\label{table_main1}
\end{table*}

\begin{table*}[t]
\small
\centering
\caption{\textbf{Evaluation on GenImage.} Accuracy (\%).
The best and second-best results are shown in \textbf{bold} and \underline{underline}, respectively.}
\begin{tabular}{lccccccccc}
\toprule
\textbf{Method} &
\textbf{Midjourney} &
\textbf{SD~1.4} &
\textbf{SD~1.5} &
\textbf{ADM} &
\textbf{GLIDE} &
\textbf{Wukong} &
\textbf{VQDM} &
\textbf{BigGAN} &z
\textbf{Avg.} \\
\midrule
ResNet50~\cite{DBLP:conf/cvpr/HeZRS16}  & 54.9 & \underline{99.7}  & 99.7 & 53.5 & 61.9 & 98.2 & 56.6 & 52.0 & 72.1 \\
DeiT-S~\cite{DBLP:conf/icml/TouvronCDMSJ21}    & 55.6  & \textbf{99.9} & \underline{99.8} & 49.8 & 58.1 & 98.9 & 56.9 & 53.5 & 71.6 \\
Swin-T~\cite{DBLP:conf/iccv/LiuL00W0LG21}    & 62.1  & \textbf{99.9} & \underline{99.8} & 49.8 & 67.6 & \underline{99.1} & 62.3 & 57.6 & 74.8 \\
CNNSpot~\cite{DBLP:conf/cvpr/WangW0OE20}   & 52.8 & 96.3 & 95.9 & 50.1 & 39.8 & 78.6 & 53.4 & 46.8 & 64.2 \\
Spec~\cite{DBLP:journals/corr/abs-1907-06515}      & 52.0 & 99.4 & 99.2 & 49.7 & 49.8 & 94.8 & 55.6 & 49.8 & 68.8 \\
F3Net~\cite{DBLP:conf/eccv/QianYSCS20}     & 50.1 & \textbf{99.9} & \textbf{99.9} & 49.9 & 50.0 & \textbf{99.9} & 49.9 & 49.9 & 68.7 \\
GramNet~\cite{DBLP:conf/cvpr/LiuQT20}   & 54.2 & 99.2 & 99.1 & 50.3 & 54.6 & 98.9 & 50.8 & 51.7 & 69.9 \\
UnivFD~\cite{DBLP:conf/cvpr/OjhaLL23}    & 73.2 & 84.2 & 84.0 & 55.2 & 76.9 & 75.6 & 56.9 & 80.3 & 73.3 \\
PatchCraft~\cite{DBLP:journals/corr/abs-2311-12397} &79.0 & 89.5 & 89.3 &  77.3 & 78.4 & 89.3&  83.7 & 72.4 &  82.3 \\
AIDE~\cite{DBLP:conf/iclr/YanLCHJ0X25}       &\underline{79.4} &\underline{99.7} &\underline{99.8} &\underline{78.5} &\underline{91.8} &98.7 &80.3 &66.9 &86.9 \\

\midrule

Ours (Training free)  &72.3 & 97.5 & 89.7 &75.9 & 86.7 &95.2 &\textbf{97.6} & \underline{90.7} &
\underline{88.2}\\

Ours (Linear probe) &\textbf{85.7} &98.1 &92.9 &\textbf{90.7} &\textbf{92.0} &93.6 &\underline{93.5} &\textbf{94.4} &\textbf{92.6} \\
\bottomrule
\end{tabular}
\label{table_main2}
\end{table*}

\begin{table*}[t]
\caption{
Comparison on the AIGCDetectionBenchmark~\citep{DBLP:journals/corr/abs-2311-12397}. Accuracy (\%) across detectors and generator categories. 
All methods are fine-tuned on ProGAN, except for DIRE-D, which follows its official setup and is trained using ADM-generated fake images.}
\begin{center}
\vspace{-3mm}
\resizebox{\textwidth}{!}{
\begin{tabular}{lcccccccccccccccccc}
\toprule
Method 
& ProGAN & StyleGAN & BigGAN & CycleGAN & StarGAN & GauGAN 
& StyleGAN2 & WFIR & ADM & Glide & Midjourney & SD v1.4 & SD v1.5 
& VQDM & Wukong & DALLE2 & \textit{Mean} \\
\midrule

CNNSpot~\cite{DBLP:conf/cvpr/WangW0OE20} & \textbf{100.00} & 90.17 & 71.17 & 87.62 & 94.60 & 81.42 & 86.91 & 91.65 
& 60.39 & 58.07 & 51.39 & 50.57 & 50.53 & 56.46 & 51.03 & 50.45 & 70.78 \\

FreDect~\cite{DBLP:journals/corr/abs-2003-08685} & 99.36 & 78.02 & 81.97 & 78.77 & 94.62 & 80.57 & 66.19 & 50.75 
& 63.42 & 54.13 & 45.87 & 38.79 & 39.21 & 77.80 & 40.30 & 34.70 & 64.03 \\

Fusing~\cite{DBLP:conf/icip/JuJKXNL22} & \textbf{100.00} & 85.20 & 77.40 & 87.00 & 97.00 & 77.00 & 83.30 & 66.80 
& 49.00 & 57.20 & 52.20 & 51.00 & 51.40 & 55.10 & 51.70 & 52.80 & 68.38 \\

LNP~\cite{DBLP:journals/corr/abs-2311-00962} & 99.67 & 91.75 & 77.75 & 84.10 & \underline{99.92} & 75.39 & 94.64 & 70.85
& 84.73 & 80.52 & 65.55 & 85.55 & 85.67 & 74.46 & 82.06 & 88.75 & 83.84 \\

LGrad~\cite{DBLP:conf/cvpr/Tan0WGW23} & 99.83 & 91.08 & 85.62 & 86.94 & 99.27 & 78.46 & 85.32 & 55.70
& 67.15 & 66.11 & 65.35 & 63.02 & 63.67 & 72.99 & 59.55 & 65.45 & 75.34 \\

UniverFD~\cite{DBLP:conf/cvpr/OjhaLL23} & 99.81 & 84.93 & 95.08 & 98.33 & 95.75 & 99.47 & 74.96 & 86.90 
& 66.87 & 62.46 & 56.13 & 63.66 & 63.49 & 85.31 & 70.93 & 50.75 & 78.43 \\

DIRE-G~\cite{DBLP:conf/iccv/WangBZWHCL23} & 95.19 & 83.03 & 70.12 & 74.19 & 95.47 & 67.79 & 75.31 & 58.05 
& 75.78 & 71.75 & 58.01 & 49.74 & 49.83 & 53.68 & 54.46 & 66.48 & 68.68 \\

DIRE-D~\cite{DBLP:conf/iccv/WangBZWHCL23} & 52.75 & 51.31 & 49.70 & 49.58 & 46.72 & 51.23 & 51.72 & 53.30 
& \textbf{98.25} & \underline{92.42} & \underline{89.45} & 91.24 & 91.63 & 91.90 & 90.90 & \underline{92.45} & 71.53 \\

PatchCraft~\cite{DBLP:journals/corr/abs-2311-12397} & \textbf{100.00} & 92.77 & \underline{95.80} & 70.17 & \textbf{99.97} & 71.58 & 89.55 & 85.80
& 82.17 & 83.79 & \textbf{90.12} & \textbf{95.38} & \textbf{95.30} & 88.91 & 91.07 & \textbf{96.60} & 89.31 \\

NPR~\cite{DBLP:conf/cvpr/TanLZWGLW24} & 99.79 & \underline{97.70} & 84.35 & \underline{96.10} & 99.35 & 82.50 & \textbf{98.38} & 65.80
& 69.69 & 78.36 & 77.85 & 78.63 & 78.89 & 78.13 & 76.11 & 64.90 & 82.91 \\

AIDE~\cite{DBLP:conf/iclr/YanLCHJ0X25}
& \underline{99.99} & \textbf{99.64} & 83.95 & \textbf{98.48} & 99.91 & 73.25 & \underline{98.00} & \underline{94.20}   
& \underline{93.43} & \textbf{95.09} & 77.20 & 93.00 & \underline{92.85} & \underline{95.16}   
& \underline{93.55} & \textbf{96.60} & \textbf{92.77} \\
\midrule
Ours (Training free) & 99.25 & 77.88 & 75.73 & 69.84 & 50.10 & \underline{99.79} & 75.71 & 77.20 & 59.90 & 75.10 & 50.60 & 74.90 & 72.90 & 94.40     
& 80.00   & 60.50    & 78.99    \\
Ours (Linear probe) & 99.59 & 89.18 & \textbf{99.18} & 89.53 & 94.52 & \textbf{99.80} & 90.92 & \textbf{98.20} & 85.76 & 93.04 & 77.32 & \underline{93.52} & 92.66 & \textbf{98.72} & \textbf{95.60} & 77.00 & \underline{92.45}   \\
\bottomrule
\end{tabular}
}
\end{center}
\label{table_main3}
\end{table*}

\subsection{Motivation}
\label{subsec:motivation}
The analyses in Sec.~\ref{subsec:token_analysis} show that DINOv3’s cross-generator 
robustness is driven by globally coherent low-frequency structure distributed across patch tokens. 
Tab.~\ref{table:tokens} further indicates that patch tokens collectively provide the strongest 
discriminative signal, but it remains unclear whether their contributions are uniform or whether 
certain tokens play a disproportionately important role in encoding authenticity cues.

This motivates a sparsity perspective: global coherence may be concentrated more strongly 
in a subset of patch tokens rather than evenly distributed. Similar sparsity phenomena 
have been observed broadly in ViTs, where only a small fraction of tokens carry most of 
the semantic or task-relevant information~\cite{DBLP:conf/aaai/XuZZSLDZXS22,DBLP:journals/corr/abs-2106-11297,DBLP:conf/nips/RaoZLLZH21}. 
If so, we expect:
\textbf{(P1)} a compact subset ($K \ll 196$) to match or exceed full-patch performance;
\textbf{(P2)} selected tokens to exhibit stronger real/fake discrimination;
\textbf{(P3)} selection patterns to remain stable across unseen generators.
These predictions call for a principled way to identify which tokens are informative.

\subsection{Rationale for Token Selection}
\label{subsec:method_rationale}

Section~\ref{subsec:motivation} indicates that authenticity cues, although global in nature, 
are not uniformly represented across patch tokens. This creates the need for a mechanism that 
identifies the tokens that most reliably capture low-frequency coherence without modifying 
the DINOv3 backbone. Such a mechanism should operate directly in the frozen feature space, 
require minimal supervision, and downweight noisy or weakly informative tokens.

To meet these requirements, This motivates the search for a simple and quantitative criterion that reflects how 
strongly each token separates real and synthetic samples. We therefore introduce 
\textbf{Fisher-Guided Token Selection (FGTS)}, a training-free procedure that scores patch 
tokens by their discriminative stability across real and synthetic distributions. FGTS 
operationalizes the sparsity perspective by selecting the tokens most reflective of the 
coherence signals that underpin cross-generator detection.

\subsection{The FGTS Framework}
\label{subsec:method_framework}

FGTS consists of two stages: an \textbf{offline token-ranking stage} that scores each patch token by its discriminative reliability, and a \textbf{lightweight inference stage} that selects and aggregates a compact subset of tokens for final prediction.

\subsubsection{Stage 1: Offline Token Ranking}

Given a small, balanced reference set (1000 real and 1000 generated images),
we compute for each patch token a scalar score indicating how well its features separate the two classes.
Specifically, we adopt the \textbf{Fisher Discriminability Ratio}, defined for the $i$-th token as:
\begin{equation}
F_i = \frac{(\mu_{\text{real}, i} - \mu_{\text{fake}, i})^2}
{\sigma^2_{\text{real}, i} + \sigma^2_{\text{fake}, i}},
\end{equation}
where $\mu_{\text{real}, i}$ and $\sigma^2_{\text{real}, i}$ denote 
the mean and variance of token $i$ for real samples, and likewise for fake ones.

A higher $F_i$ indicates that token $i$ produces features with greater between-class separation and lower within-class variability.
We rank all tokens according to $F_i$ and store the sorted index list $\mathcal{I} = [i_1, i_2, \dots, i_N]$ for downstream use.
This ranking step requires no gradient updates, is computationally inexpensive, and provides a quantitative estimate of each token’s discriminative stability.

\subsubsection{Stage 2: Lightweight Inference}

At inference time, FGTS uses only the top-$K$ ranked tokens to form a compact representation.
Given an input image:

\begin{enumerate}
    \item \textbf{Extract Features:} 
    Pass the image through a frozen backbone (for example, DINOv3) 
    to obtain $N$ patch tokens $\{p_1, \dots, p_N\}$.

    \item \textbf{Select Tokens:} 
    Retrieve the subset corresponding to the top-$K$ indices 
    $\mathcal{I}_K = \{i_1, \dots, i_K\}$.

    \item \textbf{Aggregate:} 
    Average the $K$ selected vectors to obtain an embedding 
    $z_{\text{out}} \in \mathbb{R}^D$.
\end{enumerate}

This yields a compact representation that preserves the most informative portion of the frozen feature space.

\subsection{Classification Protocols}

\noindent \textbf{(a) Training-Free Protocol.}  
No learnable parameters are introduced.  
We precompute the centroids of real and fake embeddings, $\mu_{\text{real}}$ and $\mu_{\text{fake}}$,  
using the same top-$K$ tokens from the reference set.  
A test image is classified by its cosine similarity to these centroids.  
This directly measures how separable the frozen representation is.

\noindent \textbf{(b) Linear Probe Protocol.}  
To assess linear separability under minimal supervision,  
we train a lightweight logistic regression classifier on the compact representation $z_{\text{out}}$ using only \textbf{1k real and 1k fake images from a single generator},  
while keeping the backbone entirely frozen.

\section{Experiments}
\label{sec:exp}

\subsection{Experimental Setting}
\label{subsec:setting}

\textbf{Baseline Detectors.}
We compare FGTS with two families of detectors.
(1) Traditional detectors, including CNNSpot~\cite{DBLP:conf/cvpr/WangW0OE20}, FreDect~\cite{DBLP:journals/corr/abs-2003-08685}, 
LNP~\cite{DBLP:journals/corr/abs-2311-00962}, 
Fusing~\cite{DBLP:conf/icip/JuJKXNL22},
LGrad~\cite{DBLP:conf/cvpr/Tan0WGW23},
Spec~\cite{DBLP:journals/corr/abs-1907-06515},
F3Net~\cite{DBLP:conf/eccv/QianYSCS20},
DIRE~\cite{DBLP:conf/iccv/WangBZWHCL23},
GramNet~\cite{DBLP:conf/cvpr/LiuQT20},
PatchCraft~\cite{DBLP:journals/corr/abs-2311-12397}, and
NPR~\cite{DBLP:conf/cvpr/TanLZWGLW24}.
(2) Foundation model based detectors, including
AIDE~\cite{DBLP:conf/iclr/YanLCHJ0X25},
UniverFD~\cite{DBLP:conf/cvpr/OjhaLL23},
CoDE~\cite{DBLP:conf/eccv/BaraldiCCBNC24}, and Community Forensics~\cite{DBLP:conf/cvpr/ParkO25}.

\noindent
\textbf{Datasets.}
FGTS is evaluated on three benchmarks, each following its standard reference set:
(I) So-Fake-OOD~\cite{DBLP:journals/corr/abs-2505-18660} uses $1{,}000$ real and $1{,}000$ fake images from \textbf{LDM};
(II) GenImage~\cite{DBLP:conf/nips/ZhuCYHLLT0H023} uses $1{,}000$ real and $1{,}000$ fake images from \textbf{Stable Diffusion 1.4};
and (III) AIGCDetectionBenchmark~\cite{DBLP:journals/corr/abs-2311-12397} uses $1{,}000$ real and $1{,}000$ fake images from \textbf{ProGAN}~\cite{DBLP:conf/iclr/KarrasALL18}. These reference sets are shared across both protocols.

\noindent
\textbf{Evaluation Protocols.}
All experiments are conducted using a frozen \textbf{DINOv3-ViT-7B} backbone.
Two evaluation settings are employed. 
In the \textbf{training-free} setting, we compute real and fake centroids from a 
balanced reference set of 1,000 real and 1,000 fake images and classify each test
sample by the cosine similarity between its FGTS embedding and these centroids.
In the \textbf{linear-probe} setting, a lightweight logistic regression classifier 
is trained on the same reference set while the backbone remains frozen.
Unless stated otherwise, all main results use $K=10$ selected patch tokens, 
as supported by the ablation in Sec.~\ref{subsec:ablation}.

\subsection{Experimental Results on So-Fake-OOD}
\label{subsec:sofakeppd}

We evaluate FGTS on the \textbf{So-Fake-OOD} benchmark~\cite{DBLP:journals/corr/abs-2505-18660}, 
which includes ten commercial diffusion models unseen by all detectors. 
Tab.~\ref{table_main1} reports results for both our training-free and 1k-sample linear probe variants. 
FGTS already performs competitively in the training-free setting, 
and the linear probe further improves accuracy to \textbf{87.53\%}, outperforming all comparison methods on average.

\subsection{Experimental Results on GenImage}
\label{subsec:genimage}
Tab.~\ref{table_main2} reports results on GenImage~\cite{DBLP:conf/nips/ZhuCYHLLT0H023}, 
covering eight representative generators. FGTS achieves the highest average accuracy with the linear probe (\textbf{92.6\%}), and its training-free variant ranks second, indicating strong cross-generator transfer even without learning.

\begin{table}[t]
\centering
\caption{Robustness under perturbations on So-Fake-OOD.}
\label{table:robustness}
\resizebox{0.6\linewidth}{!}{
\begin{tabular}{lccc}
\toprule
\textbf{Perturbation} & \textbf{Acc} & \textbf{AUC} & \textbf{AP} \\
\midrule
Clean          & 0.8750 & 0.9527 & 0.9561 \\
Gaussian (5)   & 0.7354 & 0.8411 & 0.8641 \\
Gaussian (10)  & 0.5531 & 0.4779 & 0.5161 \\
JPEG (70)      & 0.6981 & 0.7656 & 0.8154 \\
JPEG (80)      & 0.7514 & 0.8612 & 0.8713 \\
Resize (0.5)   & 0.8407 & 0.9225 & 0.9393 \\
Resize (0.75)  & 0.8534 & 0.9349 & 0.9411 \\
\bottomrule
\end{tabular}
}
\end{table}

\subsection{ Results on AIGCDetectionBenchmark.}
\label{subsec:CNNDetction}
We also evaluate FGTS on the AIGCDetectionBenchmark~\cite{DBLP:journals/corr/abs-2311-12397}. Results are summarized in 
Tab.~\ref{table_main3}. Despite relying only on the frozen DINOv3 backbone and a small 1k real / 1k fake reference set, FGTS remains highly competitive. The linear probe achieves an average accuracy of \textbf{92.45\%}, which is on par with the best-performing methods and clearly exceeds all traditional and most foundation model-based detectors.

\subsection{Robustness Experiment}
\label{subsec:robustness}
We evaluate the robustness of FGTS under common perturbations including Gaussian noise, JPEG compression, and image downsampling. As summarized in Tab.~\ref{table:robustness}, FGTS remains stable under mild corruptions (JPEG 80, resize 0.75).

\subsection{Ablation Experiment}
\label{subsec:ablation}

\begin{figure}[t]
    \centering
    \includegraphics[width=1.0\linewidth]{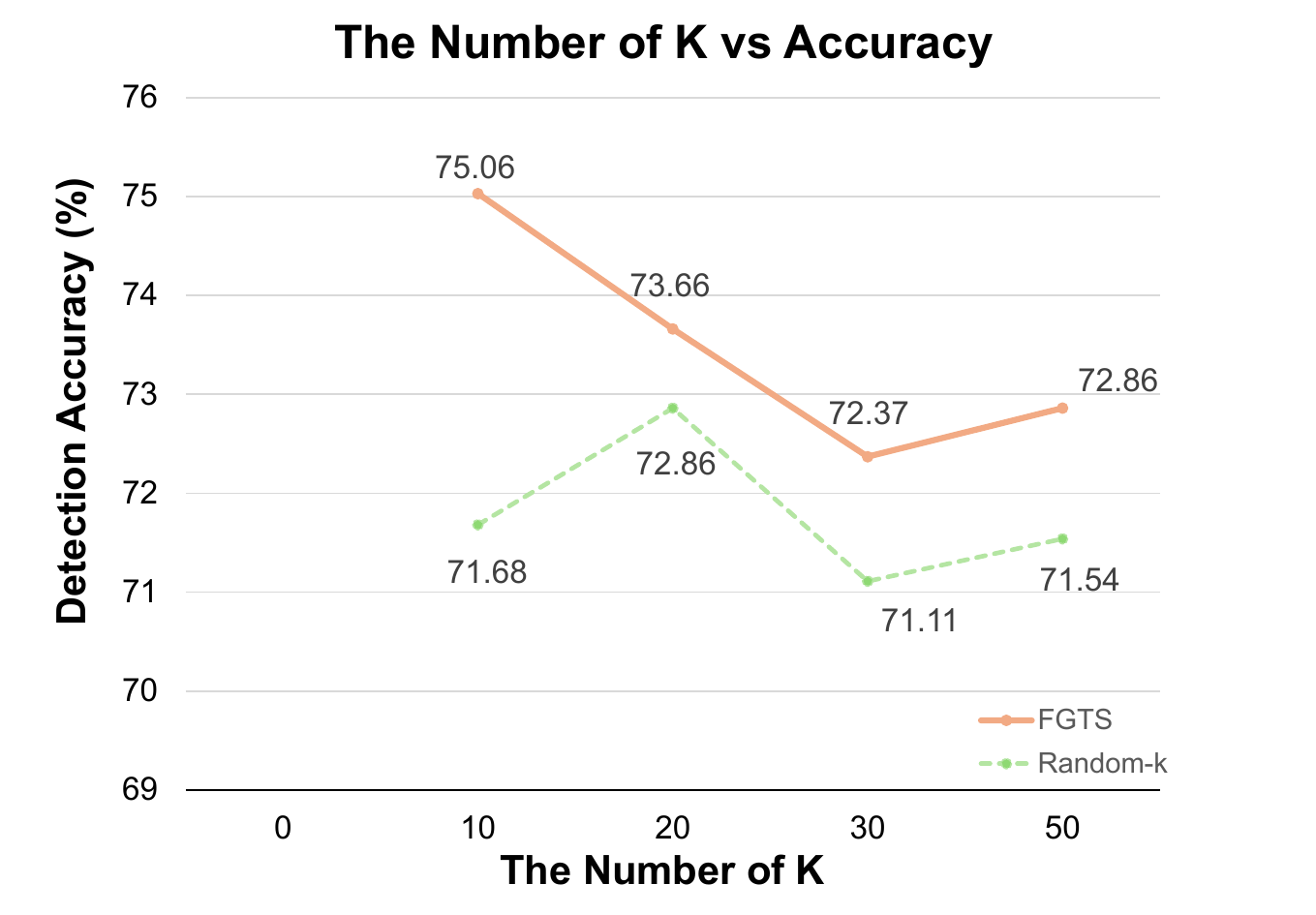}
    \caption{Impact of the number of selected tokens.}
    \label{figure:topk}
\end{figure}
\textbf{Top-$K$ Token Selection.} We assess the effect of selecting different numbers of tokens by comparing FGTS 
with a Random-$K$ baseline. For each $K \in \{10, 20, 30, 50\}$, FGTS selects 
the top-$K$ patch tokens ranked by Fisher scores, while Random-$K$ samples 
tokens uniformly at random. 
Both use the frozen DINOv3 backbone and the same 
training free protocol. As shown in Fig.~\ref{figure:topk}, FGTS consistently 
outperforms random selection across all $K$, with the largest gain at smaller 
token budgets (for example, $K=10$). This indicates that FGTS identifies a more informative subset of patch tokens.
\begin{wraptable}{r}{0.45\linewidth}
\vspace{-8pt}
\centering
\caption{Reference generators used for linear probe training on So-Fake-OOD.}
\label{table:reference}
\begin{tabular}{lc}
\toprule
Generator & Acc (\%) \\
\midrule
LDM       & 87.5 \\
SD v1.4    & 86.8 \\
ProGAN    & 84.9 \\
\bottomrule
\end{tabular}
\end{wraptable}
\noindent\textbf{Reference Generator.}
We evaluate how the choice of reference generator affects performance.
As shown in Tab.~\ref{table:reference}, using LDM, SD~v1.4, or ProGAN yields
comparable accuracy, indicating that FGTS is largely insensitive to reference.

\section{Related Work}
\label{sec:related_work}

\textbf{Image Forgery Detection.} Early detectors rely on CNNs that learn generator-specific 
artifacts~\cite{DBLP:conf/cvpr/WangW0OE20,DBLP:conf/mediaforensics/NatarajMMCFBR19,DBLP:conf/iccv/Yu0F19,DBLP:conf/icml/FrankESFKH20}, while frequency and gradient-based methods~\cite{DBLP:journals/corr/abs-2003-08685,DBLP:conf/nips/DzanicSW20,DBLP:conf/cvpr/DurallKK20} improve robustness but still fail on unseen generators. Model-centric approaches such as UniverFD~\cite{DBLP:conf/cvpr/OjhaLL23} and CoDE~\cite{DBLP:conf/eccv/BaraldiCCBNC24} rely on 
large backbones and extensive synthetic training. Recently, large vision-language models have 
also been explored for forgery detection, showing strong cross-generator robustness 
~\cite{DBLP:journals/corr/abs-2412-04292,li2024forgerygpt,DBLP:journals/corr/abs-2503-24267,DBLP:journals/corr/abs-2503-24267,DBLP:journals/corr/abs-2410-02761}. 
In contrast, FGTS avoids task-specific training entirely and uses the global features of a frozen 
DINOv3 to identify a compact subset of tokens for cross-generator detection.

\textbf{Visual Foundation Models.}
\label{subsec:visual_foundation_models}
Large visual foundation models such as CLIP~\cite{DBLP:conf/icml/RadfordKHRGASAM21,wu2023clipself,sun2023eva}, MAE~\cite{DBLP:conf/cvpr/HeCXLDG22,DBLP:conf/cvpr/00050XWYF22}, SLIP~\cite{DBLP:conf/eccv/MuK0X22}, BEiT~\cite{DBLP:conf/iclr/Bao0PW22}, and iBOT~\cite{DBLP:journals/corr/abs-2111-07832} provide strong transferable features, but their training objectives emphasize semantic alignment or reconstruction and therefore tend to pay less attention to the global low-frequency structures that matter for authenticity analysis. Self-distilled models such as DINOv2~\cite{DBLP:journals/tmlr/OquabDMVSKFHMEA24} begin to exhibit stronger global invariance, yet their representations remain less cleanly separated across token types. DINOv3~\cite{DBLP:journals/corr/abs-2508-10104}, trained at a significantly larger scale and equipped with native register tokens, exhibits noticeably cleaner global feature organization and robust invariances that align naturally with the demands of cross-generator forgery detection. 

\section{Conclusion}
\label{sec:con_limit}

We investigated how DINOv3 encodes real versus synthetic content and found that global low-frequency structure emerges as a transferable \textbf{authenticity cue}, distributed across a subset of spatial tokens rather than concentrated in non-spatial representations. Leveraging these observations, we proposed \textbf{FGTS}, a training-free token selection strategy that activates a frozen DINOv3 as a universal detector. Our results suggest that representation-centered approaches can provide a lightweight and effective alternative to task-specific training.
\textbf{Limitations and future work.}
This study provides an initial analysis and does not yet offer a complete understanding of global representations in foundation models. Moreover, the analysis centers on DINOv3, and it remains unclear how well the observations transfer to other foundation models or multimodal architectures. In addition, our investigation is limited to still images 
and does not address temporal consistency or video-specific artifacts. Finally, our evaluation covers only existing generators, which may evolve rapidly. Future work will therefore explore broader model families, temporal modeling, and more general token selection strategies.

\section{Appendix}
\label{suppl:appendix}
\appendix
\textbf{Contents of the Appendices:}

\noindent\textbf{Section~\ref{sec:impl}}: Implementation Details and Computational Efficiency.\\
\textbf{Section~\ref{sec:scaling}}: Impact of Model Scaling and Architecture.\\
\textbf{Section~\ref{sec:ablation}}: Effectiveness of Fisher-Guided Token Selection (FGTS).\\
\textbf{Section~\ref{sec:large_scale}}: Analysis of Large-Scale Adaptation.\\
\textbf{Section~\ref{sec:analy_across}}: Analysis of Performance Variations Across Generators.\\


\section{Implementation Details and Computational Efficiency}
\label{sec:impl}

This section provides the full experimental configurations used in all evaluations, followed by a detailed comparison of computational efficiency across baseline detectors.

\begin{table*}[ht]
    \centering
    \caption{\textbf{Computational Cost Comparison.} 
    For our Training free variant, the 2k images are used only to construct the reference set and do not involve parameter updates. 
    ``-'' in the GPU setup column indicates that the original paper provides no hardware details. 
    Speedup is measured relative to CNNSpot.}

    \label{tab:efficiency}
    \resizebox{\linewidth}{!}{
    \begin{tabular}{lcccccc}
        \toprule
        \textbf{Method} 
        & \textbf{Backbone} 
        & \textbf{Training Data} 
        & \textbf{Trainable Params} 
        & \textbf{Training Time} 
        & \textbf{GPU Setup} 
        & \textbf{Speedup} \\
        \midrule
        CNNSpot~\cite{DBLP:conf/cvpr/WangW0OE20}   
            & ResNet-50   
            & $\sim$720k 
            & $\sim$25M   
            & $\sim$24h 
            & -- 
            & $1\times$ \\
        UniverFD~\cite{DBLP:conf/cvpr/OjhaLL23} 
            & CLIP-ViT-L 
            & $\sim$720k 
            & $\sim$1.5k 
            & $\sim$12h 
            & -- 
            & $2\times$  \\
        CoDE~\cite{DBLP:conf/eccv/BaraldiCCBNC24}  
            & CLIP-ViT-T  
            & $\sim$12M 
            & $\sim$5M  
            & $\sim$48h 
            & RTX6000$\times$4 
            & $0.5\times$ \\
        Community-Forensics~\cite{DBLP:conf/cvpr/ParkO25} 
            & CLIP-ViT-L  
            & $\sim$5.4M 
            & $\sim$22M  
            & $\sim$72h 
            & -- 
            & $\sim0.3\times$ \\
        AIDE~\cite{DBLP:conf/iclr/YanLCHJ0X25}      
            & CLIP-ConvNeXt      
            & $\sim$300k 
            & $\sim$10M   
            & $\sim$2h 
            & A100$\times$8 
            & $12\times$ \\
        \midrule
        \textbf{Ours (Linear probe)} 
            & DINOv3-ViT-7B 
            &  2k
            & 8{,}194
            & $<$5 min 
            & RTX5090$\times$1 
            & $\sim$300$\times$ \\
        \textbf{Ours (Training free)}     
            & DINOv3-ViT-7B 
            & 2k 
            & 0    
            & 0   
            & RTX5090$\times$1     
            & N/A\\
        \bottomrule
    \end{tabular}
    }
\end{table*}

\begin{table*}[t]
\small
\centering
\setlength{\tabcolsep}{3pt}
\renewcommand{\arraystretch}{1}
\caption{\textbf{Impact of model scaling on cross-generator generalization.}
We evaluate five DINOv3 and four DINOv2 models of increasing capacity on the So-Fake-OOD benchmark. 
All numbers report accuracy (\%). Larger DINOv3 models exhibit clear scaling behavior,
while DINOv2 shows weaker improvements under the same evaluation protocol.}
\resizebox{\linewidth}{!}{
\begin{tabular}{cccccccccccc}
\toprule
\textbf{Model} &
Flux.1\_pro &
GPT-4o &
HiDream &
Ideogram~2 &
Ideogram~3 &
Imagen~3 &
Imagen~4 &
Recraft-v3 &
Seedream~3.0 &
Nano~Banana &
\textbf{Avg-acc} \\ 
\midrule
\rowcolor{gray!12}
\multicolumn{12}{c}{\textbf{DINOv3}} \\
\midrule
DINOv3-S/16 &55.6  &67.1  &69.9  &54.2  &74.4  &60.5  &68.5  &70.3  &58.3  &67.0  &64.6  \\
DINOv3-B/16 &62.3  &76.0  &81.0  &51.4  &79.7  &72.8  &70.8  &70.3  &64.3  &74.5  & 70.3 \\
DINOv3-L/16 & 71.9  &82.5  &86.6  &58.1  &88.9  &76.6  &77.7  &73.2  &68.2  &81.8 &76.7  \\
DINOv3-H/16 &76.5  &85.0  &87.7  &61.2  &88.3  &78.5  &73.7  & 72.6  &69.9  &84.7 &77.8  \\
DINOv3-7B & \textbf{79.9} & \textbf{96.0} & \textbf{96.0} &  \textbf{63.8} &  \textbf{95.9} &  \textbf{92.8} & \textbf{90.5} &\textbf{86.5} &  \textbf{77.8} &  \textbf{95.8} &  \textbf{87.5} \\
\midrule
\rowcolor{gray!12}
\multicolumn{12}{c}{\textbf{DINOv2}} \\
\midrule
DINOv2-S/14 &53.8  &50.8  &69.1  &52.8  &71.8  &53.4  &52.4  &62.6  &45.3  &51.7
&56.4  \\
DINOv2-B/14 &59.5  &54.5  &66.6  &49.7  &67.9  &55.4  &55.6  &60.2  &44.1  &57.6 & 57.1 \\
DINOv2-L/14 &61.1  &60.6  &68.5  &52.5  &68.6  &61.9  &58.4  &63.7  &57.3  &59.6 &61.2  \\
DINOv2-H/14  &62.3  &58.1  &69.4  &55.1  &66.3  &60.8  &59.8  &65.1  &59.6  &58.1 &61.4  \\

\bottomrule
\end{tabular}}
\label{tab:dinoseries}
\end{table*}

\subsection{Experimental Settings}
\label{sub:supplesetting}

\textbf{Backbone Configuration.}
All experiments employ the \textbf{DINOv3-ViT-7B}~\cite{DBLP:journals/corr/abs-2508-10104} model as a frozen feature extractor. Images are resized to $224\times224$ and normalized following the official preprocessing pipeline. Features are taken from the final transformer block without any form of test-time augmentation, fine-tuning, or distillation.

\noindent\textbf{Reference Set Construction.}
To ensure a fair comparison with fully supervised detectors, we match the exact domain assumptions adopted in prior works.
Most supervised baselines are trained using a single synthetic source per benchmark; therefore, we construct a compact reference set of \textbf{1,000 real} and \textbf{1,000 synthetic} images that mirrors their training setup:

\begin{itemize}[leftmargin=0.5cm]
\item \textbf{Real:} Following CNNSpot~\cite{DBLP:conf/cvpr/WangW0OE20}, we sample from one LSUN~\cite{DBLP:journals/corr/YuZSSX15} category (e.g., \texttt{car}).
\item \textbf{Synthetic:} For each benchmark, we select the same generator family that supervised baselines predominantly train on:
\begin{itemize}[leftmargin=0.5cm]
\item AIGCDetectionBenchmark: \textbf{ProGAN}~\cite{DBLP:conf/iclr/KarrasALL18}.
\item So-Fake-OOD: \textbf{Latent Diffusion (LDM)}~\cite{DBLP:conf/cvpr/RombachBLEO22}.
\item GenImage: \textbf{Stable Diffusion v1.5}.
\end{itemize}
\end{itemize}

This alignment removes domain discrepancies and ensures that our linear probe is evaluated under the same generative-source assumptions as the supervised detectors.

\noindent\textbf{Linear Probe.}
For settings requiring supervision, we train a minimal linear classifier on top of frozen DINOv3 features:
\begin{itemize}[leftmargin=0.5cm]
    \item \textbf{Architecture:} A single fully connected layer ($4096\rightarrow2$).
    \item \textbf{Optimization:} Adam (lr=$1\times10^{-2}$), cross-entropy loss.
    \item \textbf{Schedule:} 50 epochs, batch size 32.  
\end{itemize}
Since the backbone is frozen, training is extremely lightweight; the dominant cost of our pipeline lies in a one-time feature extraction pass over the 2k reference images. 

\subsection{Computational Efficiency}
\label{sec:compute}

A major advantage of our framework is the elimination of backbone fine-tuning, which constitutes the bulk of training cost in existing supervised detectors. We compare with widely used baselines using different backbone architectures: \textbf{CNNSpot}~\cite{DBLP:conf/cvpr/WangW0OE20} (ResNet-50~\cite{DBLP:conf/cvpr/HeZRS16}), \textbf{UniverFD}~\cite{DBLP:conf/cvpr/OjhaLL23} (CLIP-ViT-L~\cite{DBLP:conf/icml/RadfordKHRGASAM21}), \textbf{CoDE}~\cite{DBLP:conf/eccv/BaraldiCCBNC24} (CLIP-ViT-T), \textbf{Community-Forensics}~\cite{DBLP:conf/cvpr/ParkO25} (CLIP-ViT-L), and \textbf{AIDE}~\cite{DBLP:conf/iclr/YanLCHJ0X25} (CLIP-ConvNeXt~\cite{DBLP:conf/cvpr/0003MWFDX22}).

\noindent\textbf{Cost Components.}
Our compute overhead consists of:
\begin{itemize}[leftmargin=0.5cm]
    \item \textbf{Feature Extraction:} One forward pass over the 2,000-image reference set.
    \item \textbf{Linear Probe Training:} Optimization of 8,194 parameters in a single FC layer.
\end{itemize}

Tab.~\ref{tab:efficiency} compares the computational requirements of our method with prior detectors. The baselines differ substantially in both data scale and training cost: 
CNNSpot and UniverFD require optimization over large-scale datasets ($\sim$720k images), while CoDE and Community-Forensics introduce even heavier end-to-end training pipelines 
(5--12M images and tens of millions of trainable parameters). AIDE uses a ConvNeXt-based OpenCLIP backbone trained on 8$\times$A100 GPUs and remains considerably more expensive than any frozen-feature approach.

In contrast, our method optimizes only a single linear layer with 8,194 parameters, 
and its dominant computational cost is a one-time feature extraction pass over the 2k reference images (approximately 3 minutes on a single RTX 5090). 
The training free variant removes this step entirely by performing no parameter updates. 
Consequently, our linear probe reduces the training cost by over \textbf{300$\times$} relative to CNNSpot, while maintaining competitive performance across all benchmarks.

\section{Impact of Model Scaling and Architecture}
\label{sec:scaling}

In this section, we systematically examine how model scale and architectural design affect cross-generator generalization. We evaluate a broad range of capacities within the DINO family, including \textbf{DINOv3: ViT-S/B/L/H-16}~\cite{DBLP:journals/corr/abs-2508-10104} and \textbf{DINOv2: ViT-S/B/L/H-14}~\cite{DBLP:journals/tmlr/OquabDMVSKFHMEA24}, to assess whether larger backbones consistently yield stronger robustness under our evaluation protocol.

These two model families offer a natural architectural comparison: DINOv3 adopts a more recent self-supervised training strategy and scaling recipe, while DINOv2 represents an earlier generation of feature learning. Evaluating both across multiple scales allows us to analyze how much cross-generator performance is driven by model capacity versus architectural and training differences.

All experiments in this section follow a unified evaluation protocol: a lightweight linear probe is trained on the LDM reference set (1k real and 1k fake), and performance is evaluated on the So-Fake-OOD benchmark using the proposed FGTS representation.

\begin{figure}[th]
    \centering
    \includegraphics[width=1.0\linewidth]{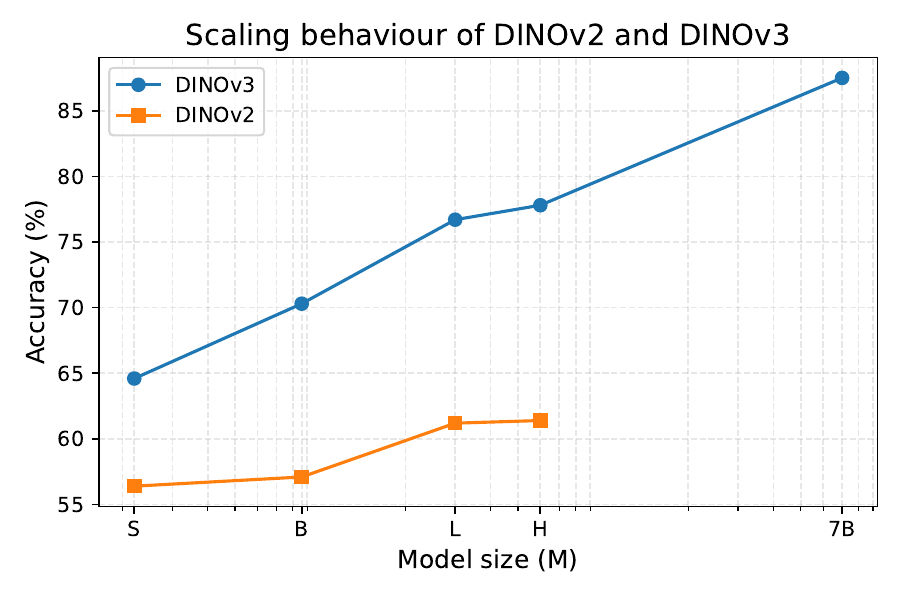}
    \caption{\textbf{Scaling behavior of DINOv2 and DINOv3 on cross-generator detection.}
Accuracy on So-Fake-OOD is plotted against model size (millions of parameters). 
DINOv3 exhibits a clear scaling trend from ViT-S/16 to ViT-7B, while DINOv2 shows limited 
improvement with increased capacity.}
\vspace{-5mm}
    \label{figure:scale}
\end{figure}

The results in Tab.~\ref{tab:dinoseries} and Fig.~\ref{figure:scale} show that DINOv3 exhibits a clear and consistent scaling trend. Accuracy increases steadily from ViT-S/16 to ViT-7B, indicating that both model capacity and the updated training strategy contribute to stronger cross-generator robustness. In comparison, DINOv2 shows only mild gains from S/14 to H/14, and its overall performance remains noticeably lower than DINOv3 at similar scales. This suggests that capacity alone is not sufficient without the architectural and training improvements present in DINOv3.

Given the smooth upward trajectory of DINOv3, particularly the substantial improvement at the 7B scale, it is reasonable to expect that further increasing \textbf{model capacity}, together with training strategies that reinforce global consistency, may continue to enhance cross-generator generalization.

\section{Effectiveness of Fisher-Guided Token Selection (FGTS)}
\label{sec:ablation}

In this section, we evaluate the effectiveness of the proposed FGTS method under  training free setting.
We compare four DINOv3 models across three inference configurations: (1) using all tokens, (2) using only patch tokens, and (3) using the FGTS-selected token subset. 
This comparison allows us to isolate the contribution of token selection and to 
assess how much discriminative information FGTS preserves relative to full token configuration.

\begin{figure}[t]
    \centering
    \includegraphics[width=1.0\linewidth]{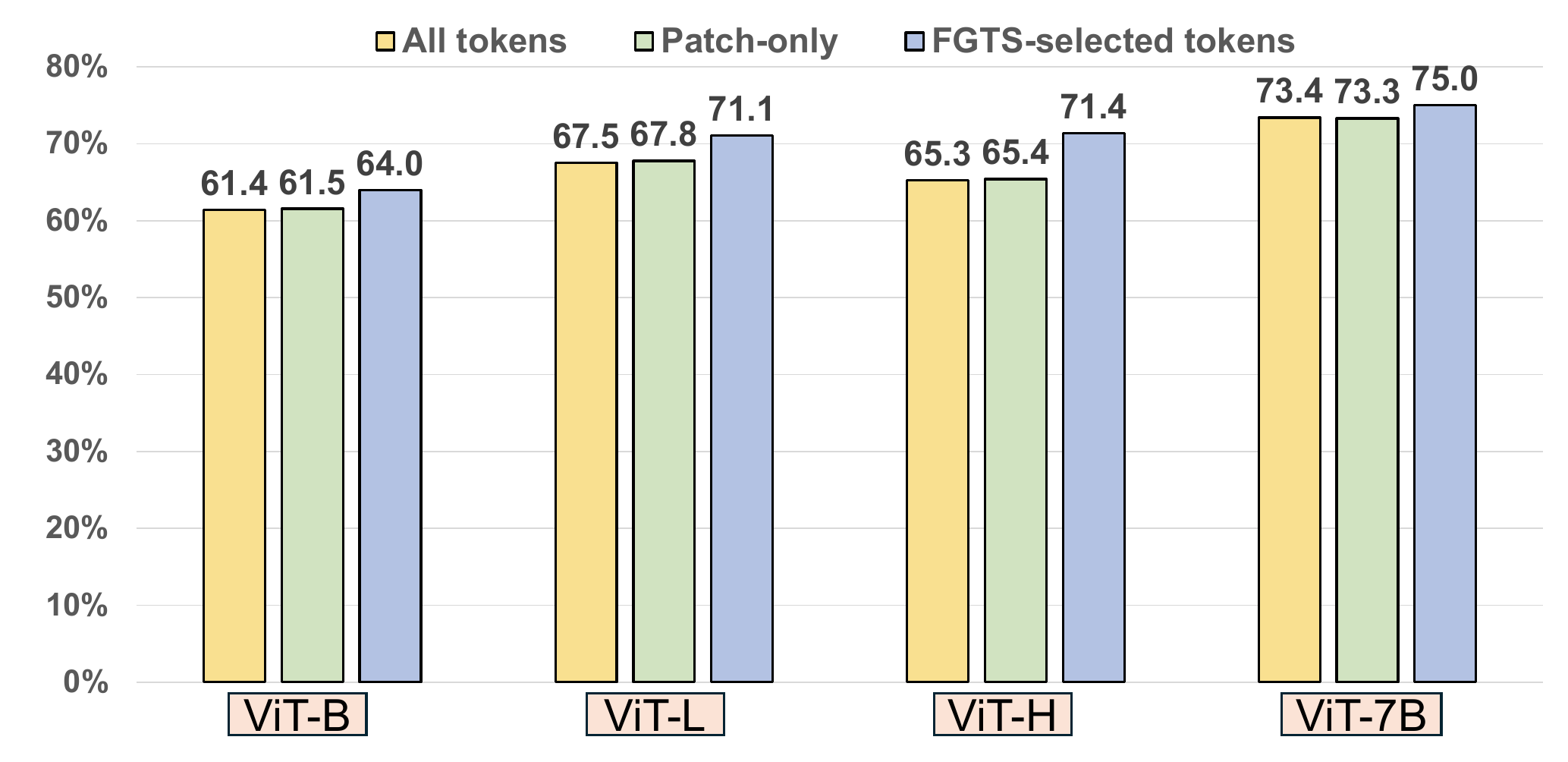}
    \caption{\textbf{Effectiveness of FGTS across DINOv3 scales.}
    FGTS consistently outperforms the patch-only baseline and approaches 
    all-token accuracy across ViT-B/16 to ViT-7B.}
    \label{figure:FGTS}
\end{figure}

As illustrated in Fig.~\ref{figure:FGTS}, FGTS consistently outperforms both the patch-only baseline and the all-token approach across all DINOv3 scales. Notably, FGTS achieves substantial improvements over using all tokens: +2.6\% on ViT-B, +3.6\% on ViT-L, and a striking +6.1\% on ViT-H. The performance gap narrows to +1.6\% on ViT-7B, suggesting that the largest models encode authenticity cues with sufficient redundancy that even naive token aggregation becomes effective. For mid-scale models, however, FGTS plays a critical role in filtering out noisy or less discriminative tokens that would otherwise dilute the authenticity signal. These results demonstrate that FGTS provides a more effective representation than naive token aggregation, with particularly significant benefits at mid-to-large scales.

\begin{figure}[th]
    \centering
    \includegraphics[width=1.0\linewidth]{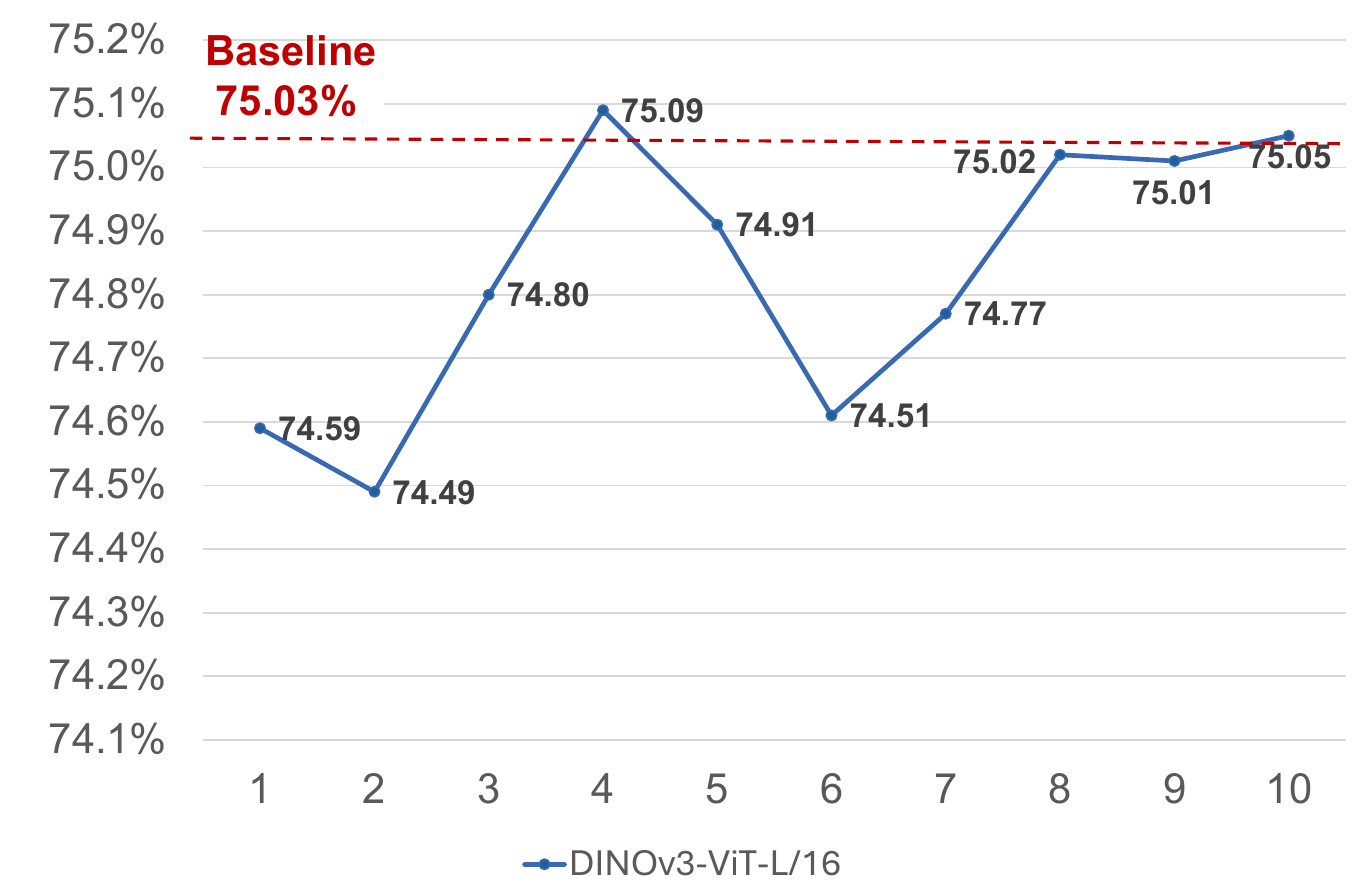}
    \caption{\textbf{Effect of large-scale fine-tuning on DINOv3-L/16.}
    Accuracy on So-Fake-OOD across 10 epochs fluctuates around the frozen 
    baseline, indicating that supervised adaptation fails to strengthen 
    cross-generator representations.}
    \label{figure:degrade}
\end{figure}

\begin{figure*}[th]
    \centering
    \includegraphics[width=1.0\linewidth]{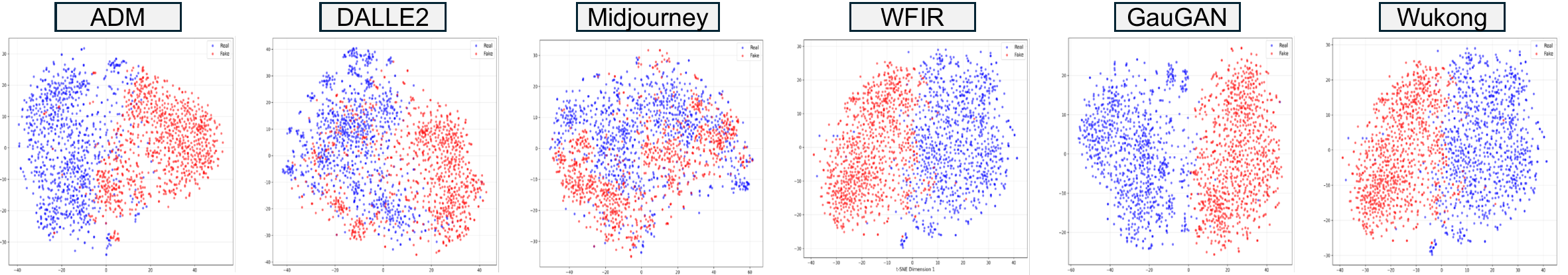}
    \caption{\textbf{t-SNE visualization of feature distributions} (Blue: Real, Red: Fake). The \textbf{first three} generators (ADM, DALLE-2, Midjourney) exhibit significant feature entanglement, which explains the performance gap compared to fine-tuned methods. In contrast, the \textbf{last three} (WFIR, GauGAN, Wukong) show clear linear separability, validating the effectiveness of our frozen backbone in capturing structural artifacts.}
    \label{figure:tsne}
\end{figure*}

\section{Analysis of Large-Scale Adaptation}
\label{sec:large_scale}
To further understand the limitations of large-scale adaptation, we analyze how heavy supervised fine-tuning affects the intrinsic representation quality of DINOv3. We examine whether large-scale fine-tuning on diverse training categories can enhance cross-generator generalization. To investigate this systematically, we follow the data construction protocol of UniverFD~\cite{DBLP:conf/cvpr/OjhaLL23}, which uses 20 semantic categories such as \emph{airplane}, \emph{car}, and \emph{dog}. For each category, we collect the corresponding real images and generate an equal number of synthetic images using LDM~\cite{DBLP:conf/cvpr/RombachBLEO22}. This mirrors the traditional deepfake training setup employed by prior work, except that we use LDM rather than ProGAN to generate synthetic images. In total, the fine-tuning dataset contains 360k real images and 360k LDM generated images.

The model is fine-tuned for 10 epochs, and all other training configurations follow the settings described in Section~\ref{sub:supplesetting}. We evaluate the resulting representations using a linear probe on top of DINOv3-L/16 and report performance on the So-Fake-OOD benchmark. This setup directly measures how large-scale supervised adaptation affects the model's ability to retain global authenticity cues.

As shown in Fig.~\ref{figure:degrade}, large-scale supervised fine-tuning fails to improve the cross-generator generalization of DINOv3-L/16. Across 10 epochs of training, accuracy on So-Fake-OOD fluctuates around the frozen baseline (75.03\%), ranging from 74.49\% to 75.09\%. The best checkpoint at epoch 4 achieves 75.09\%, showing only marginal improvement over the baseline, while several epochs exhibit slight performance drops, indicating instability during adaptation.

This suggests that the global authenticity cues encoded during DINOv3 pretraining are already highly effective and difficult to enhance through standard supervised learning. The lack of consistent improvement, combined with training instability, indicates that large-scale fine-tuning offers little benefit for cross-generator generalization in this setting. In contrast, the lightweight linear probe achieves comparable performance without the computational cost and potential risks of full backbone adaptation.

\section{Analysis of Performance Variations Across Generators}
\label{sec:analy_across}

\begin{figure}[th]
    \centering
    \includegraphics[width=1.0\linewidth]{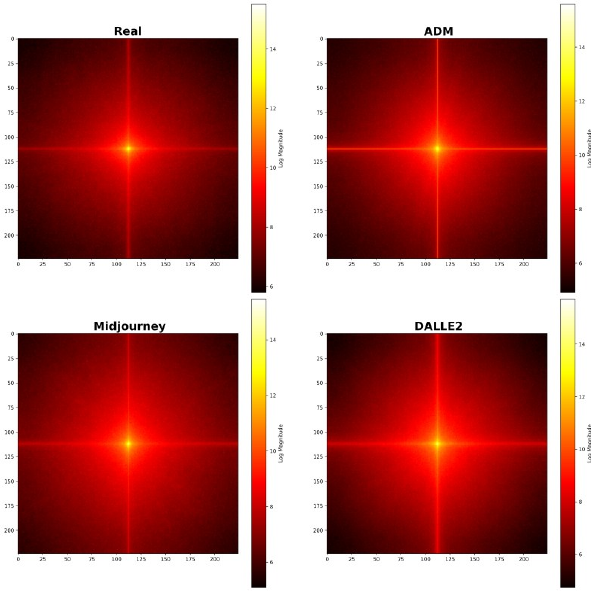}
    \caption{\textbf{Frequency spectrum analysis.} ADM, DALLE-2, and Midjourney exhibit frequency distributions nearly identical to real images, with energy concentrated in low-frequency components (central bright cross). This high-fidelity low-frequency replication explains the feature entanglement observed in Fig.~\ref{figure:tsne} and the detection challenges for frozen low-frequency-biased representations.}
    \label{figure:frequency}
\end{figure}

While FGTS demonstrates strong overall generalization, its performance varies across generators in AIGCDetectionBenchmark~\citep{DBLP:journals/corr/abs-2311-12397}. We achieve over 95\% accuracy on certain generators (e.g., WFIR, VQDM) but observe reduced performance on Midjourney (77.32\%) and DALLE-2 (77.00\%). We investigate the underlying causes through feature space and frequency domain analysis.

\noindent\textbf{t-SNE Visualization.} Fig.~\ref{figure:tsne} visualizes frozen DINOv3 features via t-SNE for six generators. The results reveal two distinct patterns: (1) \textbf{Challenging generators} (ADM, DALLE-2, Midjourney) exhibit significant real/fake feature entanglement, making linear separation difficult. (2) \textbf{Separable generators} (WFIR, GauGAN, Wukong) show clear linear separability with distinct real/fake clusters. This pattern directly correlates with their respective detection accuracy.

\noindent\textbf{Frequency Domain Analysis.} Fig.~\ref{figure:frequency} shows frequency spectra for Real, ADM, Midjourney, and DALLE-2. All four exhibit comparable low-frequency energy concentration (central cross pattern) with minimal high-frequency content. Notably, challenging generators (ADM, DALLE-2, Midjourney) show closer spectral similarity to real images than separable generators. This observation is consistent with our finding that DINOv3 relies predominantly on low-frequency global structure: when generators produce low-frequency patterns similar to real images, frozen features exhibit reduced discriminability.

Although FGTS achieves the best cross-dataset generalization overall, the above analysis reveals room for further improvement on these generators that successfully replicate low-frequency characteristics of real images.

{
    \small
    \bibliographystyle{ieeenat_fullname}
    \bibliography{main}
}


\end{document}